\documentclass[conference]{IEEEtran}
\usepackage[utf8]{inputenc}
\usepackage[T1]{fontenc}

\usepackage[english]{babel}
\usepackage{xcolor}
\usepackage{hyperref}
\hypersetup{pdfborder={0 0 0},colorlinks=true,urlcolor=blue,linkcolor=blue,citecolor=blue,bookmarks=true}
\hypersetup{pdftitle={Extreme Value Modelling of Feature Residuals for Anomaly Detection in Dynamic Graphs}}
\hypersetup{pdfauthor={Sevvandi Kandanaarachchi, Conrad Sanderson, Rob J. Hyndman}}

\usepackage{caption}
\DeclareCaptionLabelFormat{abbreviated}{\textbf{Fig.~{#2}}}
\usepackage[shortlabels]{enumitem}
\usepackage{newtxtext}   
\usepackage{inconsolata} 
\usepackage{graphicx}
\graphicspath{{figures/},{Figures/}}

\usepackage{amsfonts}  
\usepackage{amsmath}
\usepackage{mathrsfs}  
\usepackage{bm}

\usepackage[absolute]{textpos}

\begin{document}

\title
  {
  \normalsize
  \scalebox{1.7}{\normalsize\textbf{Extreme Value Modelling of Feature Residuals}}\\
  \scalebox{1.7}{\normalsize\textbf{for Anomaly Detection in Dynamic Graphs}}
  }

\author
  {
  Sevvandi Kandanaarachchi\textsuperscript{{\tiny~}$\dagger$},
  Conrad Sanderson\textsuperscript{{\tiny~}$\dagger\ddagger$},
  Rob J. Hyndman\textsuperscript{{\tiny~}$\diamond$}\\
  ~\\
  \textsuperscript{$\dagger$}{\tiny~}\textit{Data61{\tiny~}/{\tiny~}CSIRO, Australia;}~
  \textsuperscript{$\ddagger$}{\tiny~}\textit{Griffith University, Australia;}~
  \textsuperscript{$\diamond$}{\tiny~}\textit{Monash University, Australia}
  \vspace{-1ex}
  }

\maketitle

\begin{abstract}

Detecting anomalies in a temporal sequence of graphs can be applied is areas
such as the detection of accidents in transport networks and cyber attacks in computer networks.
Existing methods for detecting abnormal graphs can suffer from multiple limitations,
such as high false positive rates
as well as difficulties with handling variable-sized graphs and non-trivial temporal dynamics.
To address this, 
we propose a technique where temporal dependencies are explicitly modelled via time series analysis
of a large set of pertinent graph features,
followed by using residuals to remove the dependencies.
Extreme Value Theory is then used to robustly model and classify any remaining extremes,
aiming to produce low false positives rates.
Comparative evaluations on a multitude of graph instances show that the proposed approach obtains considerably better accuracy than
\textit{TensorSplat} and \textit{Laplacian Anomaly Detection}.
\end{abstract}


\begin{textblock}{13.44}(1.28,14.80)
\hrule
\vspace{1ex}
\noindent
\scalebox{0.78}{\textbf{{$^\ast$}~Published in:}  International Conference on Soft Computing \& Machine Intelligence (ISCMI), pp.~32--37, 2024. DOI:~\href{https://doi.org/10.1109/ISCMI63661.2024.10851659}{\tt 10.1109/ISCMI63661.2024.10851659}}
\end{textblock}

\renewcommand{\baselinestretch}{0.970}\small\normalsize

\section{Introduction}

\begin{figure}[!tb]
  \centering
  \begin{minipage}{1\columnwidth}\
    \centering
    \begin{minipage}{0.04\textwidth}
      \footnotesize
      \textbf{(a)}
    \end{minipage}
    \hfill
    \begin{minipage}{0.92\textwidth}
      {\includegraphics[width=\textwidth,height=0.3\textwidth]{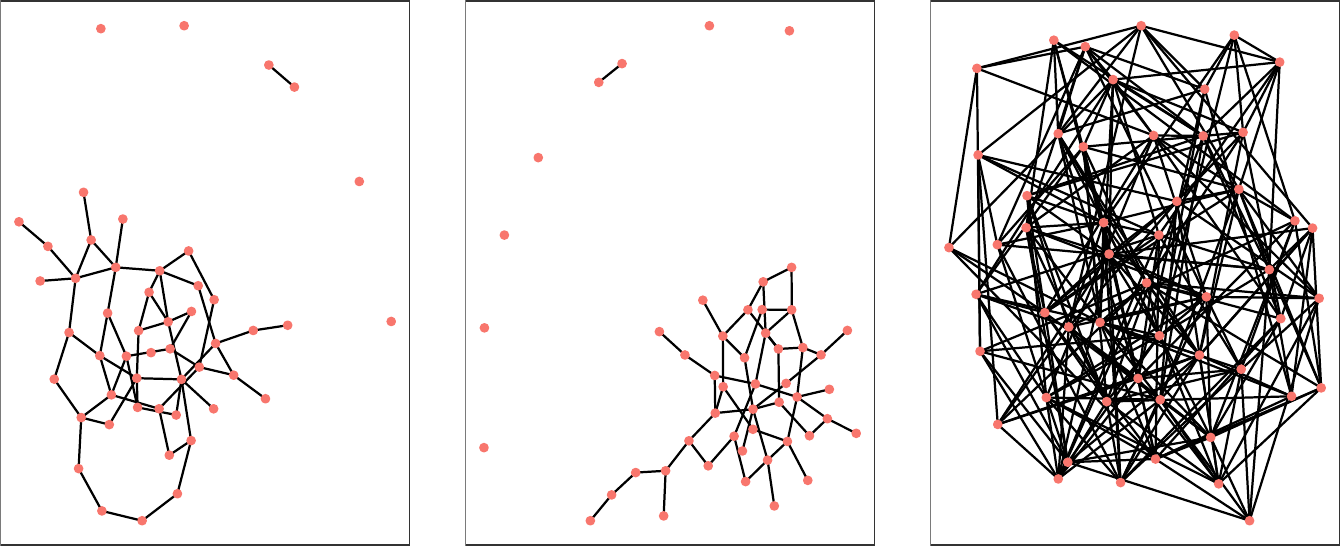}}
    \end{minipage}
  \end{minipage}
  \vspace{1ex}
  \hrule
  \vspace{1ex}
  \begin{minipage}{1\columnwidth}\
    \centering
    \begin{minipage}{0.04\textwidth}
      \footnotesize
      \textbf{(b)}
    \end{minipage}
    \hfill
    \begin{minipage}{0.92\textwidth}
      {\includegraphics[width=\textwidth,height=0.3\textwidth]{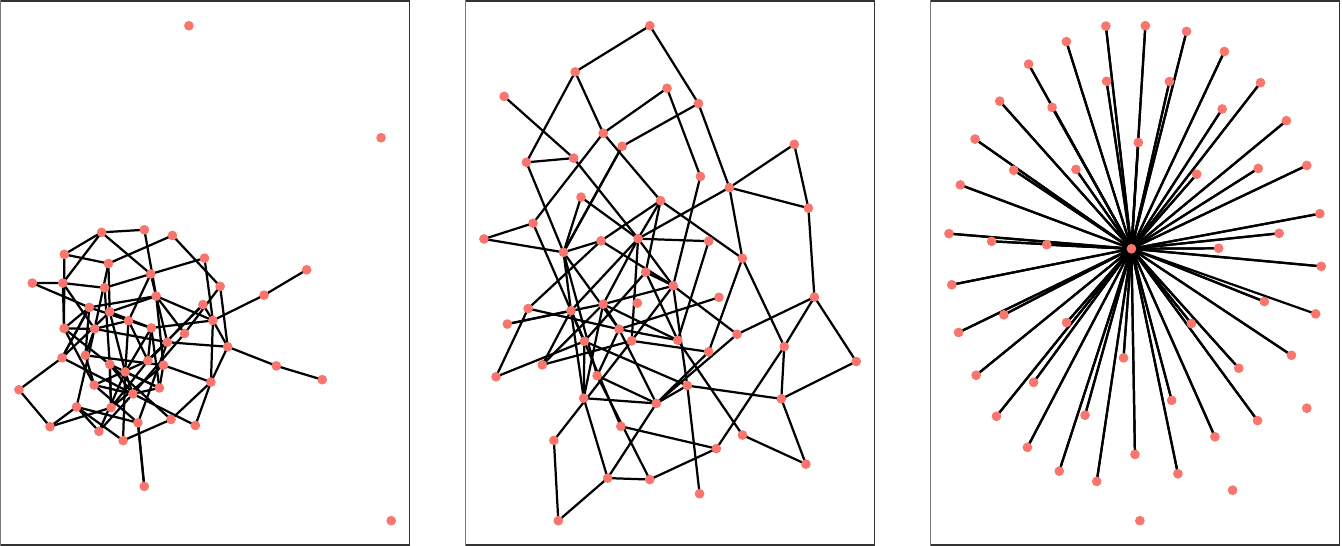}}
    \end{minipage}
  \end{minipage}
  \caption
    {
    Two examples of graph sequences with anomalies, with each sequence ordered left-to-right.
    \textbf{(a)}~Graphs generated using the Erd\H{o}s-R\'enyi random graph model~\cite{frieze2015introduction};
    the first two graphs have an edge probability of 0.05,
    while the last graph is abnormal with an edge probability 0.20.
    \textbf{(b)}~Graphs with 100 edges;
    the first two graphs have randomly selected edges,
    while the last graph is abnormal as all its edges are connected to a central vertex.
    }
  \label{fig:example_anomalies}
\end{figure}

Many real-world networks such as transport, electricity, and social networks
can be represented as graphs,
which model the relationships between sets of entities.
An~ordered sequence of such graphs can be used to model temporal changes,
such as the evolution of relationships as well as the addition and removal of entities~\cite{Akoglu2015,Ekle_2024,Liu9599560,Ma2021}.
We refer to sequences of graphs as \textit{dynamic graphs}.
Detecting the time points at which the sequence is anomalous
can be useful for detecting (or possibly averting) harmful events,
such as blackouts in electricity networks and cyber attacks in computer networks.
Example graph sequences with anomalies are shown in Fig.~\ref{fig:example_anomalies}. 

In contrast to detecting anomalous vertices in a graph~\cite{liu2022bond},
in this work we focus on graph-level anomaly detection,
which is determining whether an entire graph is abnormal within a sequence of graphs~\cite{Ma2021}.
Anomaly detection techniques for dynamic graphs can be categorised into four broad types,
based on:
(i)~features,
(ii)~decompositions,
(iii)~clustering,
and
(iv)~moving windows~\cite{Akoglu2015}.

In feature-based approaches, a fixed-dimensional vector representation of each graph is obtained,
and a dedicated distance function measures the similarity between successive graphs.
Decomposition-based methods use standard tensor decomposition methods to generate a summary of each graph.
Cluster-based methods concentrate on clusters of vertices (subgraphs) rather than whole graphs.
In window-based methods, graphs within a moving window are assumed to represent the range of normalcy;
a successive graph is compared with the window-derived model to determine whether it is abnormal.

An advantage of feature-based methods is that the number of vertices in graphs at various time points can differ,
while the same number of features is extracted from each graph,
thereby simplifying similarity measures between graphs.
A~possible disadvantage is that the features are empirically chosen,
and hence it can be challenging to determine a-priori which features are most useful.
Matrix decomposition-based methods omit the feature selection problem
by using the graph representation directly
(eg.~via Laplacian matrices~\cite{diestel2024graph}).
However, when the number of vertices changes, the size of the representation matrix also changes,
complicating measurements of similarity between consecutive graphs.
Hence in decomposition-based methods it is often assumed
that the number of vertices for all time points is fixed and known beforehand.
This is a major limitation, and is often addressed via ad-hoc workarounds.
Cluster-based methods may give more insights on why a graph in abnormal,
however such methods may have issues with handling community evolution aspects;
communities can naturally dissolve over time,
new communities can arise, and multiple communities can merge~\cite{Jiao_2023}.
Window-based methods indirectly take into account that graphs can change over time,
however due to unknown rates of natural change,
determining the optimal window size is haphazard.

In addition to the abovementioned limitations,
a notable shortcoming of many existing methods for anomaly detection in dynamic graphs
is the lack of generalisation to graphs with large (but natural and expected) variations at successive time points.
Existing methods are typically designed to only handle a sequence of graphs which share a similar structure,
with consecutive graphs differing by an edge or vertex~\cite{Ma2022}.

In this work we propose a feature-based anomaly detection technique
with the dual aim of handling sequences with variable-sized graphs (ie.~varying number of vertices)
and potentially large but natural changes between consecutive graphs.
To side-step the feature selection problem, we use a smorgasbord of pertinent features
that describe various aspects of graphs.
To handle complex change patterns that traditional similarity measures cannot capture,
temporal dependencies are explicitly modelled via time series analysis,
followed by using residuals (differences between expected and actual values) to remove the dependencies.
In order to avoid alert fatigue~\cite{Ban_2023},
Extreme Value Theory~\cite{Coles_2001,Reiss_2007} is then used to robustly model and classify any remaining extremes,
with the aim of producing low false positives rates.

We continue the paper as follows.
The proposed approach is described in Section~\ref{sec:proposed},
including feature extraction and temporal modelling.
Section~\ref{sec:experiments} presents an evaluation of the proposed method on four classes of graphs,
and compares its performance against two existing methods for anomaly detection.
Section~\ref{sec:conclusion} provides concluding remarks and avenues for further exploration.

\section{Proposed Approach}
\label{sec:proposed}

In a given graph, the relationships between sets of entities are referred to as edges (also known as links),
while the entities are referred to as vertices (also know as nodes).
Consider an ordered sequence of temporal graphs {\small $\{\mathcal{G}_t\}_{t = 1}^T$} where each graph is 
denoted as {\small $\mathcal{G}_t = (\mathcal{V}_t\, \mathcal{E}_t)$},
where {\small $\mathcal{V}_t$} are the vertices and {\small $\mathcal{E}_t$} are the edges.
A graph {\small $\mathcal{G}_t$} at time $t$ is considered as abnormal
if its likelihood is sufficiently low given preceding graphs in the temporal sequence
{\small $p_{t|t-1} = \mathcal{P}(\mathcal{G}_t \mid \mathcal{G}_1,\dots,\mathcal{G}_{t-1})$}.

We propose to approximate {\small $p_{t|t-1}$} by first transforming each graph (with arbitrary number of edges)
to a fixed-dimensional space via a dedicated feature extraction function {\small $f$},
followed by modelling the temporal aspects of the graph source 
as an Auto-Regressive Integrated Moving Average (ARIMA) model~\cite{HyndmanBook2013}
of the temporal sequence {\small $\left\{ f(\mathcal{G}_t) \right\}_{t = 1}^T$}.
Temporal dependencies are then removed through
differences (residuals) between the expected values predicted by ARIMA
and the observed values,
followed by projection into a low dimensional space. 
Extreme Value Theory (EVT) is then used to robustly model low density regions
that may indicate anomalies.
Each of the processing steps is elucidated below.

\subsection{Features}
\label{sec:features}

We designate the feature extraction function as \mbox{\small $f: \mathscr{G} \to \mathbb{R}^n$},
with {\small $\mathscr{G}$} indicating the space defined by a given graph source.
The {\small $n$}-dimensional feature vector representation of {\small $\mathcal{G}_t$ under~$f$}
is designated as \mbox{\small $\bm{x}_t \in \mathbb{R}^n$},
with {\small $\mathcal{F}$} indicating the associated feature space,
possibly as a manifold within Euclidean space.
If the extraction function {\small $f$} is well-designed,
the two spaces {\small $\mathscr{G}$} and {\small $\mathcal{F}$}
should possess a high degree of accord:
two graphs either close or far in space {\small $\mathscr{G}$}
should be either close or far, respectively, in space {\small $\mathcal{F}$}.

As in many applications the full details on the graph source are generally unavailable,
it is not known \mbox{a-priori} which features would be most appropriate
for adequately describing graphs for the purpose of anomaly detection.
To address this, we propose to use a rich and diverse set of pertinent features.
For each graph {\small $\mathcal{G}_t$}, a set of 20 features (scalars) is extracted, as listed below;
details on each feature can be obtained in~\cite{latora2017complex}.
For each feature that requires a distribution summary,
the 99-th percentile is used as the descriptive scalar.
The features are summarised as:

\vspace{1ex}

\begin{small}
\begin{enumerate}[{(a)},leftmargin=*]
\itemsep=1ex

\item
Number of vertices in graph $\mathcal{G}_t$.

\item
Number of edges in graph $\mathcal{G}_t$.

\item
99-th percentile of the triangle distribution of graph $\mathcal{G}_t$
(num.~of triangles associated with each vertex).

\item
99-th percentile of the degree distribution of graph $\mathcal{G}_t$
(num.~of edges associated with each vertex).

\item
Edge density, defined as the ratio of num.~of edges to num.~of all achievable edges.

\item
Transitivity, 
measuring the proportion of vertices where adjacent vertices are also connected.

\item
Assortativity coefficient,
measuring the degree of similarity in external attributes across connected vertices (homophily).
If no external attributes are available,
the degree of each vertex is used as the attribute to compute assortativity.

\item
Mean graph distance, defined as the average of all lowest path distances between vertices;
only distances of available paths are taken into account.

\item
Diameter, defined as the lowest distance between the two most distant vertices.

\item
Percentage of non-connected (isolated) vertices.

\item
Vertex connectivity, defined as the minimum num.~of vertices
that must be excised so that the graph ceases to be strongly connected.

\item
Global efficiency, defined as the mean of inverse distances between all pairs of vertices.

\item
Number of connected components in the graph,
where a connected component denotes a subgraph with vertices that are connected to each other by a sequence of edges.

\item
99-th percentile of the distribution of the num.~of vertices in each connected component. 

\item
The percentage of vertices with closeness centrality $\geq 0.8$,
where the closeness centrality of a given vertex
is measured as the reciprocal of the sum of the shortest paths to all vertices.

\item
99-th percentile of the distribution of betweenness centrality for all vertices,
where the betweenness centrality for a given vertex is the number of shortest paths that pass through it.

\item
99-th percentile of the distribution of PageRank values for all vertices~\cite{LangvilleMeyer2006}.

\item
Representation of hub scores (measure of vertex importance)
via the principal eigenvalue of $\mathbf{A}\mathbf{A}^T$,
where $\mathbf{A}$ denotes the adjacency matrix,
and the principal eigenvector denotes the hub scores of vertices.

\item
Representation of authority scores via the principal eigenvalue of $\mathbf{A}^T\mathbf{A}$,
where $\mathbf{A}$ denotes the adjacency matrix,
and the principal eigenvector denotes the authority scores of vertices.
The authority and hub score features are different for directed graphs,
and are equivalent for undirected graphs.

\item
99-th percentile of the distribution of coreness for all vertices,
where cores meassure graph community aspects.
A $k$-core is a maximal subgraph with minimum degree of at least~$k$;
each vertex in a $k$-core has degree $\geq k$.
If a given vertex belongs to a $k$-core (and not to a $(k+1)$-core),
its coreness is indicated as~$k$.

\end{enumerate}
\end{small}

\subsection{Temporal Modelling and Residuals}

\begin{figure}[!t]
  \centering
  \begin{minipage}{1\columnwidth}\
    \centering
    \begin{minipage}{0.04\textwidth}
      \footnotesize
      \textbf{(a)}
    \end{minipage}
    \hfill
    \begin{minipage}{0.92\textwidth}
      {\includegraphics[width=\textwidth,height=0.4\textwidth]{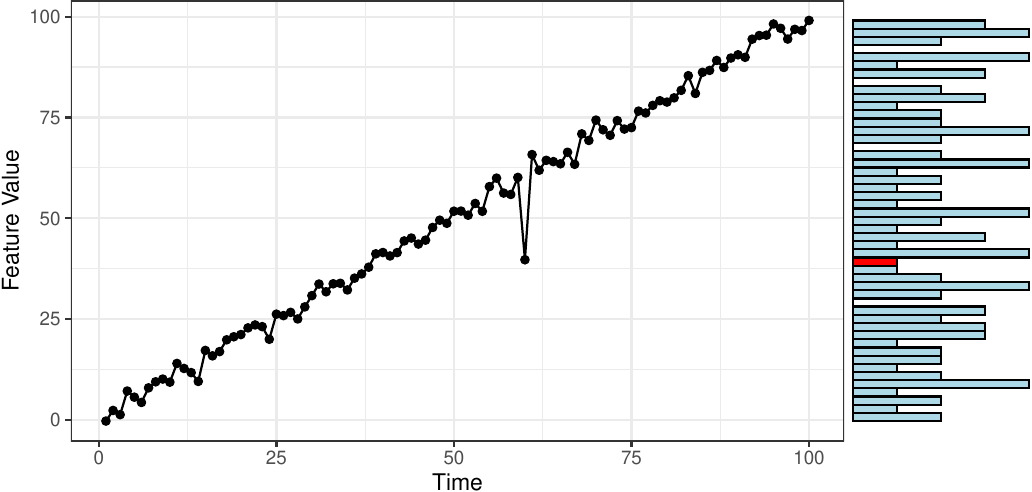}}
    \end{minipage}
  \end{minipage}
  
  \vspace{1ex}
  
  \begin{minipage}{1\columnwidth}\
    \centering
    \begin{minipage}{0.04\textwidth}
      \footnotesize
      \textbf{(b)}
    \end{minipage}
    \hfill
    \begin{minipage}{0.92\textwidth}
      {\includegraphics[width=\textwidth,height=0.4\textwidth]{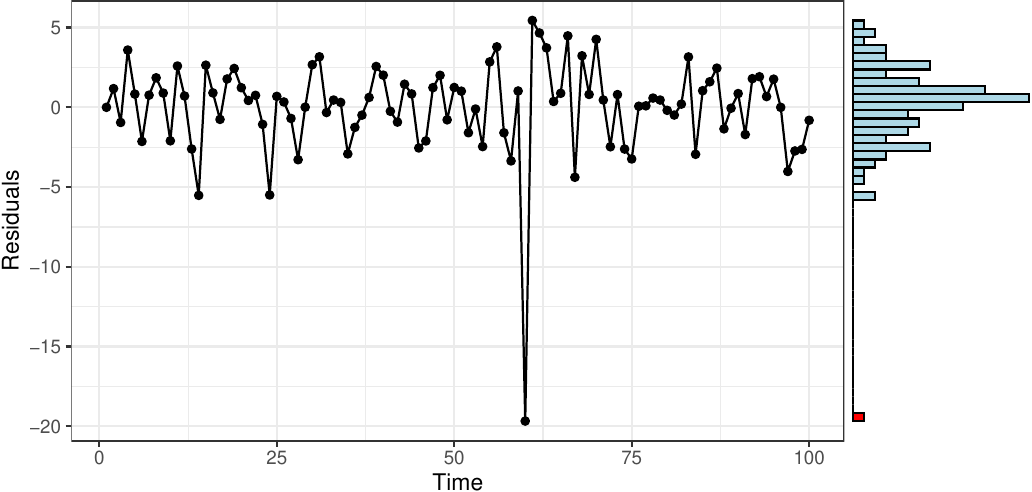}}
    \end{minipage}
  \end{minipage}
  \vspace{-1ex}
  \caption
    {
    \textbf{(a)}.
    Example of natural change in an arbitrary graph feature over time,
    with an anomaly at time~=~60.
    The distribution of the features is shown on the y-axis (right of plot), 
    with the bin corresponding to the anomaly marked in red.
    Without taking into account the temporal context,
    the anomaly is not discernible via inspecting the feature distribution.
    \textbf{(b)}.~Residuals from temporal modelling of data in~(a).
    The corresponding distribution is given on the y-axis (right of plot), 
    where the anomaly is clearly discernible.
    }
  \label{fig:feature_distribution}
\end{figure}

A given graph source can produce temporal dependencies that might be observable in the feature space.
For instance, the number of edges can increase in a given time period.
A~sudden decrease in the number of edges at a particular time point can be considered as an anomalous event.
However, without explicitly taking into account the temporal context, the decrease may not be sufficiently prominent.
A conceptual example of this phenomenon is illustrated in Fig.~\ref{fig:feature_distribution}(a), 

The ARIMA time series prediction method~\cite{HyndmanBook2013} on the graph features
is used for modelling temporal dependencies.
An ARIMA model with parameter set {\small $(p, d, q)$} for a general time series {\small $\{y_t\}$} is described via:

\vspace{-1ex}
\noindent
\begin{small}
\begin{eqnarray}
y_{ t}
\mbox{~=~}
& c & \mbox{~+~} \phi_1 y_{d, t-1}
  \mbox{~+~} \phi_2 y_{d, t-2}
  \mbox{~+~} \ldots 
  \mbox{~+~} \phi_p y_{d, t-p} \label{eqn:arima}
  \\
& ~ & \mbox{~+~} \theta_1\epsilon_{t-1}
  \mbox{~+~} \theta_2 \epsilon_{t-2}
  \mbox{~+~} \ldots
  \mbox{~+~} \theta_q \epsilon_{t-q}
  \mbox{~+~} \epsilon_{t}
  \nonumber
\end{eqnarray}
\end{small}

\noindent
where
parameter~{$d$} describes the number of differences used,
ie.~{\small $y_{d,t}$} is the {$d$}-th differenced time series,
where {\small $y_{1,t} \mbox{~=~} y_t \mbox{~--~} y_{t\mbox{-}1}$}
and {\small $y_{2,t} \mbox{~=~} y_{1,t} \mbox{~--~} y_{1, t\mbox{-}1}$} and so on;
parameter~$p$ represents the number of lagged $y_t$ values used in the model,
(ie.~the number of past observations in the time series that are used to model~$y_{t}$);
parameter~$q$ represents the number of proceeding, lagged values in the error terms used to model $y_t$.
For a given parameter set {\small $(p, d, q)$},
the variables {\small $c$}, {\small $\phi_1, ..., \phi_p$} and {\small $\theta_1, ..., \theta_q$} in Eqn.~(\ref{eqn:arima})
are estimated using maximum likelihood techniques~\cite{hyndman2008automatic}.
For a given time series,
the best suited parameters {\small $(p, d, q)$}
are estimated using state space models or step-wise selection~\cite{hyndman2008automatic}.

Given the scalar time series of the \mbox{$i$-th} graph feature,
denoted as {\small $\bm{x}_{ . ,i} \mbox{~=~} \{ x_{t, i} \}_{t = 1}^T$},
the corresponding residuals are obtained as {\small $e_{t,i} \mbox{~=~} x_{t, i} \mbox{~--~} \hat{x}_{t, i}$},
where {\small $\hat{x}_{t,i}$} is the predicted value of {\small $x_{t,i}$} (obtained from the corresponding ARIMA model),
given {\small $x_{1,i}, ..., x_{t-1,i}$}.
If~the temporal dependencies are sufficiently well-modelled,
then the residuals have no temporal dependencies
and have an approximate normal distribution~{\small $\mathcal{N}(0, \sigma^2)$}.
Fig.~\ref{fig:feature_distribution}(b) demonstrates the effect of residuals stemming from temporal modelling,
facilitating the detection of anomalies.

\subsection{Dimensionality Reduction}
\label{sec:dimreduction}

Finding anomalies in high dimensional spaces can be difficult,
as many points can be located in low density regions,
with relatively high inter-point distances~\cite{zimek2012survey}.
To partially address this and to facilitate further modelling,
we employ robust PCA~\cite{croux2007algorithms} for dimensionality reduction,
projecting scaled versions of the residuals to a 2-dimensional space.

Scaling is employed to ensure that the residuals occupy similar value ranges.
For the $i$-th feature, the residuals {\small $\bm{e}_{., i}$}
are scaled via
\mbox{\small $\bm{y}_{.,i} = {\tilde{s}_i^{~\mbox{-}1}} \cdot {\left( \bm{e}_{., i} - \tilde{\mu}_i \right) }$},
with 
{\small $\tilde{\mu}_i$} and {\small $\tilde{s}_i$}
denoting the mean and standard deviation for {\small $\bm{e}_{., i}$},
respectively, 
calculated via data in the 2.5 and 97.5 percentiles
(ie. in trimmed form).

Using the columns {\small $\bm{y}_{.,i}$} we construct the matrix {\small $\bm{Y}$} which is used for robust PCA computation.
We denote the output coordinates by {\small $\bm{Z}_{\text{full}} = \text{robust\_PCA\_mapping}(\bm{Y})$},
and consider the 2-dimensional space {\small $\bm{Z} = \{ \bm{z}_{.,1}, \bm{z}_{.,2} \}$},
with {\small $\bm{z}_{.,i}$} denoting the $i$-th ordered coordinates of {\small $\bm{Z}_{\text{full}}$}.

\subsection{Modelling via Extreme Value Theory}
\label{sec:classification}

Incidence of high false positives is a paramount challenge in anomaly detection~\cite{Ban_2023}.
When an anomaly detection system produces a large number of false positives,
the trust in the system decreases,
putting its practical applicability into question.
In contrast to simply declaring all extremes as anomalous (which can yield high false positive rates), 
EVT~\cite{Coles_2001,Reiss_2007} is used to explicitly model the extremes,
thereby allowing for robust detection of unusual extremes.

The anomalous points lie in low  density regions in the reduced dimensionality space.
By using the absolute logarithm of kernel density estimates {\small ${v}_i \mbox{~=~} |\log( \text{kde}(({z}_{i, 1}, {z}_{i,2}))|$},
points with low density are transformed to large {$v$} values,
while points with high density to small {$v$} values.
Via EVT, the extremes of $\bm{v}$ are modelled using a Generalised Pareto Distribution (GPD),
and points over a user-defined percentile are declared as anomalous.
The distribution function of GPD is defined as:

\vspace{-1ex}
\noindent
\begin{small}
\begin{equation}
   H(v) = 1 - \left( 1 + {\left(\xi v\right)}~/~{\sigma_u} \right)^{-1/\xi}
  \label{eq:POT1}
\end{equation}
\end{small}
\vspace{-3ex}

\noindent
where the domain of $H$ is {\small $\{v: v >0\, \, \text{and} \, \,  (1 + \xi v/\sigma_u) >0  \}$},
and {\small $\sigma_u = \sigma + \xi(u- \mu)$}.
The parameters $\mu$, $\sigma$ and $\xi$ denote the location, scale and shape of the distribution, respectively.
Parameter $u$ is used for selecting a subset of the given points that represent extremes,
and is typically set to the 90-th percentile~\cite{Kandan_2022}.
The probability of observing a value higher than $v_i$ is given by: 

\vspace{-2ex}
\begin{small}
\begin{equation}
P(v \geq v_i) = \int_{v_i}^{\infty} H(v) ~ dv
\end{equation}
\end{small}

\noindent
The above probability is converted into an anomaly score,
on which a user-set threshold can be applied in order to classify
a given graph as either normal or abnormal:

\begin{small}
\begin{equation}
    \text{anomaly\_score}(v_i) = 1 - P(v \geq v_i)
\end{equation}
\end{small}

The GPD is a versatile distribution and encapsulates various types of extremes specified by the shape parameter $\xi$.
For {\small $\xi \mbox{~=~} 0$} the extremes are distributed according to a Gumbel distribution;
for {\small $\xi \mbox{~<~} 0$} it is a Weibull distribution;
for {\small $\xi \mbox{~>~} 0$} it is a Fréchet distribution.
Thus, by modelling $\bm{v}$ using a GPD,  all types of extreme behaviour are encapsulated.

The values of the parameters are estimated from values {\small $v_1, \ldots, v_k$},
which are the~{\small $k$} values greater than the threshold $u$. 
Estimation is accomplished via maximising the log-likelihood defined below.
Numerical methods are used for the maximisation as analytical methods are not feasible~\cite{Coles_2001,Reiss_2007}.

For {\small $\xi \neq 0$}, the log-likelihood is derived from Eqn.~\eqref{eq:POT1} as:

\vspace{-2ex}
\begin{small}
\begin{equation}
  \mathcal{L}(\sigma, \xi) = -k\log \sigma  - (1 + 1/ \xi) \sum\nolimits_{i = 1}^k \log\left(1 + \xi v_i/\sigma \right)
  \label{eq:POT2}
\end{equation}
\end{small}

\noindent
provided {\small $(1 + \sigma^{-1}\xi v_i) \mbox{~>~} 0$} for {\small $i \in \{1, \ldots,k \}$}.
In the case {\small $\xi \to 0$} the distribution is given~by:

\vspace{-1ex}
\begin{small}
\begin{equation}
H(v) =  1 - \exp\left( - {v}/{\sigma} \right)\, , \quad v > 0
\end{equation}
\end{small}

\noindent
making the log-likelihood:

\vspace{-1ex}
\begin{small}
\begin{equation}
\mathcal{L}(\sigma) = - k \log \sigma - \sigma^{-1} \sum\nolimits_{i = 1}^k v_i
\end{equation}
\end{small}

\noindent
As the extremes are modelled using a GPD,
there are many instances in which no anomalies are identified
as the extremes are deemed to belong to the normal part of the extreme value distribution,
thereby giving low false positive rates.

\section{Comparative Evaluation}
\label{sec:experiments}

To gauge the performance of the proposed method,
we compare it against two popular methods focused on graph-level anomaly detection:
TensorSplat \cite{Koutra2012} and Laplacian Anomaly Detection (LAD) \cite{Huang2020}.
The evaluation is performed on three distinct types of graphs,
which are used as proxies for various applications.

\subsection{TensorSplat and LAD}
\label{sec:tensorsplat_and_lad}

Both TensorSplat and LAD are decomposition-based methods,
briefly summarised as follows.
Given a graph sequence {\small $\left\{ \mathcal{G}_t \right\}_{t=1}^T$},
TensorSplat first forms a 3rd order tensor with size {\small $N\times N \times T$},
where {\small $N\mbox{~=~}\max\left\{ |\mathcal{G}_t| \right\}_{t=1}^{T}$} is the maximum number of vertices across all graphs in the sequence.
To handle graph sequences where the number of vertices varies across graphs,
the adjacency matrix of each graph in the sequence is iteratively expanded by inserting zero-valued rows and columns
to account for ``missing'' vertices.
For example, if a new vertex appears at $t \mbox{~=~} 2$,
then the adjacency matrix for $t \mbox{~=~} 1$ is expanded as a way of simulating the new vertex in the earlier graph.
Hence a limitation of this approach is that earlier adjacency matrices can have more zero entries than the latter ones,
which may not be an accurate depiction of the evolving graph sequence.
The tensor is decomposed into 3 vectors
{\small $\bm{u}_1 \in \mathbb{R}^N$}, {\small $\bm{u}_2 \in \mathbb{R}^N$} and {\small $\bm{u}_3 \in \mathbb{R}^T$}.
The decomposition finds vectors {\small $\bm{u}_i$} such that {\small $\bm{u}_1 \circ \bm{u}_2 \circ \bm{u}_3 \approx X$},
where {\small $X$} denotes the tensor and $\circ$ denotes the outer product.
Vectors $\bm{u}_1$ and $\bm{u}_2$ give details about the adjacency matrix while $\bm{u}_3$ corresponds to the time component.
Anomalies are found using vector $\bm{u}_3$, and anomaly scores are given as output.
As such, TensorSplat has a further limitation as it does not directly account for temporal dependencies.

LAD computes the eigenvectors of the Laplacian matrix
and using a moving window computes the dissimilarity
between the current Laplacian eigenvector and the Laplacian eigenvector representing normal behaviour.
More specifically,
LAD uses the Laplacian matrix {\small $L_t = D_t \mbox{~--~} A_t$},
where {\small $D_t$} is the diagonal degree matrix and {\small $A_t$} is the adjacency matrix of graph {\small $\mathcal{G}_t$}.
Singular Value Decomposition (SVD) is performed on each $L_t$
to obtain the top $k$ singular values $\bm{\sigma}_t$, where $k$ is a parameter.
Using a window of size $\ell$ the matrix {\small $C = [\sigma_{t-\ell - 1}, \cdots, \sigma_{t-1}]$} is formed
and the left singular vector (with largest singular value) of {\small $C$} is computed via SVD. 
This is called the normal behaviour vector~{\small $\bar{\sigma}_t$}.
The dissimilarity between {\small $\bar{\sigma}_t$} and the current vector {\small ${\sigma}_t$} is computed,
and is called the $z$~score.
The temporal dependencies are modelled via differences between successive $z$~scores. 
Both the raw $z$ scores and differenced $z$ scores are used as anomaly scores.
A limitation of LAD is that differences between scores can only model basic temporal fluctuations.
More complex dynamics such as periodic and longer-term fluctuations are not captured by successive differences.
Furthermore, determining the appropriate window size is haphazard as it varies between applications.

\subsection{Graph Types}
\label{sec:graph_types}

\begin{figure}[!b]
  \centering
  \begin{minipage}{1\columnwidth}
    \begin{minipage}{0.29\textwidth}
      \centering
      \footnotesize
      \includegraphics[width=1\textwidth,height=1\textwidth]{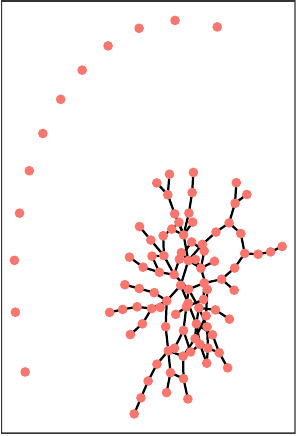}\\
      \textbf{(a)}
    \end{minipage}
    \hfill
    \begin{minipage}{0.29\textwidth}
      \centering
      \footnotesize
      \includegraphics[width=1\textwidth,height=1\textwidth]{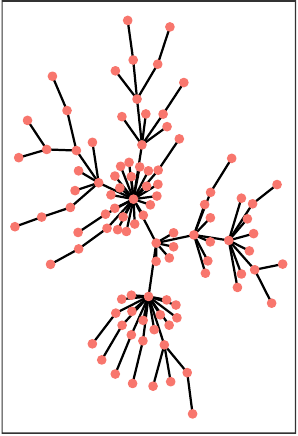}\\
      \textbf{(b)}
    \end{minipage}
    \hfill
    \begin{minipage}{0.29\textwidth}
      \centering
      \footnotesize
      \includegraphics[width=1\textwidth,height=1\textwidth]{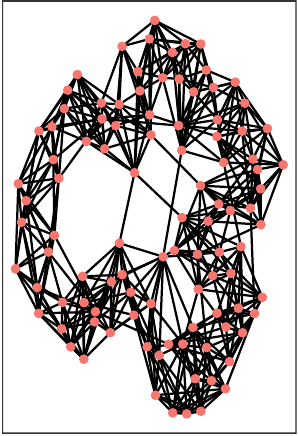}\\
      \textbf{(c)}
    \end{minipage}
  \end{minipage}
  \vspace{-1ex}
  \caption
    {
    Example graphs generated by:
    \textbf{(a)}~Erd\H{o}s-R\'enyi model~\cite{frieze2015introduction},
    \textbf{(b)}~Barab\'asi-Albert model~\cite{albert2002statistical},
    \textbf{(c)}~Watts-Strogatz model~\cite{Watts1998}.
    }
  \label{fig:graph_types}
\end{figure}

Three graph types are considered (see Fig.~\ref{fig:graph_types}):
Erd\H{o}s-R\'enyi random graph model~\cite{frieze2015introduction}, 
Barab\'asi-Albert preferential attachment model~\cite{albert2002statistical},
and Watts-Strogatz small world model~\cite{Watts1998}.

The Erd\H{o}s-R\'enyi model~\cite{frieze2015introduction}
describes graphs containing $n$~vertices with edge probability $p$,
with edges are placed randomly and independently of other edges.
The number of possible edges is~{\small ${n \choose 2}$}.
The probability of generating a graph with {\small $M$} edges is {\small $p^{M}(1-p)^{{n \choose 2} - M}$}.
Due to inherent randomness, these graphs exhibit low clustering. 
This model is often used as a starting point for other graph models. 
 
In the Barab\'asi-Albert preferential attachment model~\cite{barabasi},
new vertices are more likely to link with already well connected vertices,
in contrast to the Erd\H{o}s-R\'enyi model.
The likelihood of a new vertex connecting to vertex $i$ is proportional to {\small $k_i^{\alpha}$},
with {\small $k_i$} denoting the degree of vertex~$i$,
while exponent~$\alpha$ is a parameter indicating the degree of preferential attachment.
For {\small $\alpha \mbox{~=~} 0$}, the model degrades to a random graph.
For {\small $\alpha \mbox{~=~} 1$}, linear preferential attachment occurs.
For {\small $\alpha \in (0,1)$}
the attachment is reduced,
referred to as sub-linear preferential attachment.
Finally, for {\small $\alpha \mbox{~>~} 1$} super-linear preferential attachment is obtained.
Preferential attachment models are often employed for modelling social media networks
which typically exhibit high degrees of clustering.

The Watts-Strogatz model~\cite{Watts1998} rewires regular graphs
with the aim of introducing increasing amounts of disorder.
The rewired graphs have a disposition to be significantly clustered, while having small path lengths.
The graphs begin with a ring lattice with $n$~vertices and $k$~edges per vertex,
with each edge randomly rewired with probability~$p$.
Such graphs are exemplified in neurological systems and power generator backbones.

\subsection{Experiments}

Four experiments are performed.
In each experiment a specific graph generating model is used.
The anomaly scores of a method and ground-truth labels
are used to produce a Receiver Operator Characteristic (ROC) curve, 
which provides true positive rates and false positive rates for all possible threshold values.
The Area Under the ROC Curve (AUC) is then computed as an indicator of performance~\cite{Flach_2011}.
Higher AUC values indicate higher overall accuracy.

For each experiment there are four parameter settings.
Each setting is independently run 10 times, to account for randomness in the graphs.
This results in 10 AUC values for each parameter setting for each method.
These values are shown in summarised form via \textit{boxplots},
visually encapsulating their distribution along with the 25-th, 50-th and 75-th percentiles.

\subsubsection{Experiment 1}

A static environment is simulated via Erd\H{o}s-R\'enyi graphs, all configured with the same edge probability~{\small $p$}.
A time series of 100 graphs is generated, with each graph comprised of 100 vertices with edge probability {\small $p \mbox{~=~} 0.05$}.
An abnormal graph with edge probability~{\small $p_*$}
replaces the original graph at time {\small $t \mbox{~=~} 50$}. 
Four instances are evaluated, with each instance using a specific \mbox{\small $p_* \in \{ 0.10, 0.15, 0.20, 0.25 \}$}.
To account for randomness in graph generation, for each instance the process was repeated 10 times.

We note that in this experiment, temporal dependencies are not present;
a graph at {\small $t \mbox{~=~} 2$} is comparable to the graph at {\small $t \mbox{~=~} 1$}.
As such, finding anomalies should be relatively straightforward.

\begin{figure}[!tb]
  \centering
  \begin{minipage}{1\columnwidth}
    \begin{minipage}{0.025\textwidth}
      \centering
      \rotatebox{90}{\scriptsize AUC}
    \end{minipage}
    \begin{minipage}{0.31\textwidth}
      \centering
      {\scriptsize ~~~~~~TensorSplat}\\
      \vspace{0.1ex}
      \includegraphics[width=1\textwidth,height=1\textwidth]{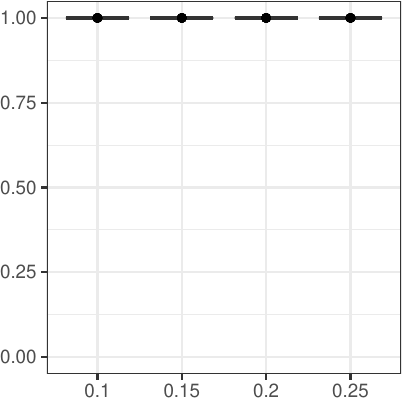}\\
      \vspace{-1ex}
      {\scriptsize ~~~~~~edge probability}
    \end{minipage}
    \hfill
    \begin{minipage}{0.31\textwidth}
      \centering
      {\scriptsize ~~~~~~LAD}\\
      \vspace{0.1ex}
      \includegraphics[width=1\textwidth,height=1\textwidth]{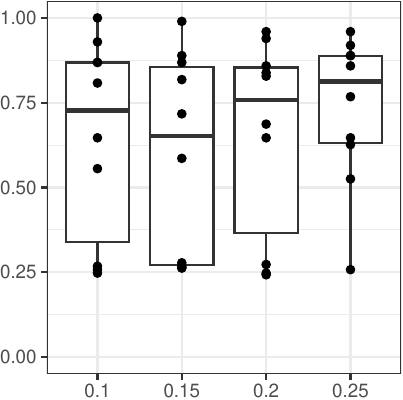}\\
      \vspace{-1ex}
      {\scriptsize ~~~~~~edge probability}
    \end{minipage}
    \hfill
    \begin{minipage}{0.31\textwidth}
      \centering
      {\scriptsize ~~~~~~\textbf{proposed}}\\
      \vspace{0.1ex}
      \includegraphics[width=1\textwidth,height=1\textwidth]{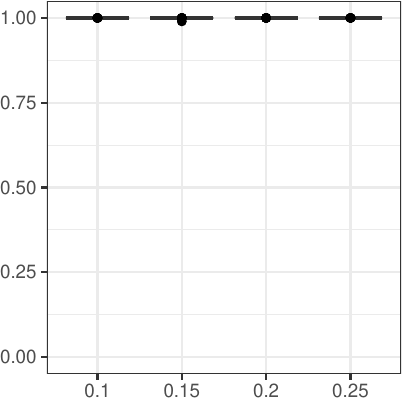}\\
      \vspace{-1ex}
      {\scriptsize ~~~~~~edge probability}
    \end{minipage}
  \end{minipage}
  \caption
    {
    Results for experiment~1.
    Performance is shown in terms of boxplots that summarise the distribution of AUC values.
    Higher AUC values indicate higher accuracy.
    The results are obtained from 10 time series of 100 graphs,
    with each graph comprised of 100 vertices with edge probability $p \mbox{~=~} 0.05$,
    generated according to the Erd\H{o}s-R\'enyi model~\cite{frieze2015introduction}.
    In each time series, an abnormal graph is placed at $t \mbox{~=~} 50$,
    with four distinct edge probabilities: $0.1$, $0.15$, $0.2$, $0.25$ (marked on x-axis).
    }
  \label{fig:experiment1}
  \vspace{0.5ex}
  \hrule  
\end{figure}

The results shown in Fig.~\ref{fig:experiment1} indicate that 
both the proposed method and TensorSplat achieve near perfect results.
The results also indicate that LAD performs poorly in all configurations of this experiment.

\subsubsection{Experiment 2}

A relatively straightforward dynamic environment is simulated via Erd\H{o}s-R\'enyi graphs,
with increasing edge probability $p$.
A time series of 100 graphs is generated, with each graph comprised of 100 vertices,
with $p$ linearly increasing from {\small $0.05$} to {\small $0.50$}.
An abnormal graph with edge probability {\small $p_\ast + 0.2727$}
replaces the original graph at {\small $t \mbox{~=~} 50$}.
Four instances are considered, with each instance using a specific {\small $p_\ast \in \{ 0.05, 0.10, 0.15, 0.20 \}$}.
The value {\small $0.2727$} is the original edge probability at the considered time point.

\begin{figure}[!tb]
  \centering
  \begin{minipage}{1\columnwidth}
    \begin{minipage}{0.025\textwidth}
      \centering
      \rotatebox{90}{\scriptsize AUC}
    \end{minipage}
    \begin{minipage}{0.31\textwidth}
      \centering
      {\scriptsize ~~~~~~TensorSplat}\\
      \vspace{0.1ex}
      \includegraphics[width=1\textwidth,height=1\textwidth]{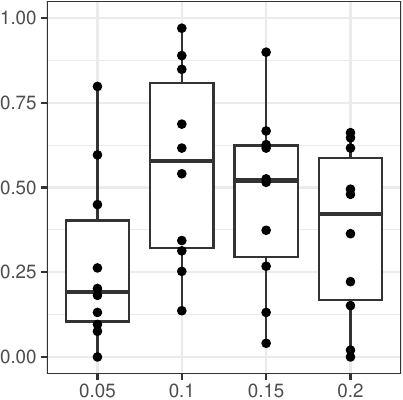}\\
      \vspace{-1ex}
      {\scriptsize ~~~~~~$p_\ast$}
    \end{minipage}
    \hfill
    \begin{minipage}{0.31\textwidth}
      \centering
      {\scriptsize ~~~~~~LAD}\\
      \vspace{0.1ex}
      \includegraphics[width=1\textwidth,height=1\textwidth]{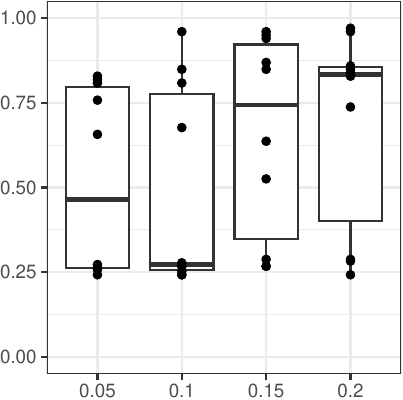}\\
      \vspace{-1ex}
      {\scriptsize ~~~~~~$p_\ast$}
    \end{minipage}
    \hfill
    \begin{minipage}{0.31\textwidth}
      \centering
      {\scriptsize ~~~~~~\textbf{proposed}}\\
      \vspace{0.1ex}
      \includegraphics[width=1\textwidth,height=1\textwidth]{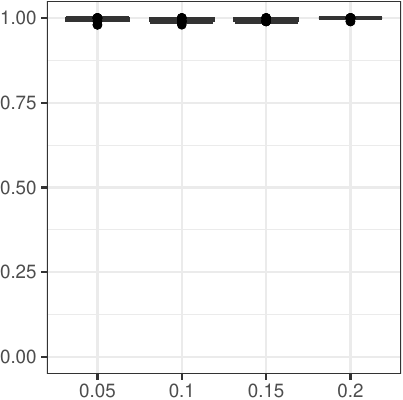}\\
      \vspace{-1ex}
      {\scriptsize ~~~~~~$p_\ast$}
    \end{minipage}
  \end{minipage}
  \vspace{-0.5ex}
  \caption
    {
    Results for experiment~2.
    As per Fig.~\ref{fig:experiment1},
    but the edge probability is linearly increasing from $0.05$ to $0.50$ in each time series of 100 graphs.
    In each time series, an abnormal graph is placed at $t \mbox{~=~} 50$,
    with edge probability $p_\ast \mbox{~+~} 0.2727$,
    where $p_\ast \in \{ 0.05, 0.10, 0.15, 0.20 \}$.
    }
  \label{fig:experiment2}
  \vspace{0.5ex}
  \hrule
\end{figure}

The results given in Fig.~\ref{fig:experiment2} indicate that the proposed method achieves near perfect performance, 
while both TensorSplat and LAD achieve notably worse performance in this more difficult setting.

\subsubsection{Experiment 3}

The Barab\'asi-Albert preferential attachment model is used to simulate a more challenging dynamic environment,
where the likelihood of a new vertex connecting to a current vertex~$i$ is proportional to {\small $k_i^{\alpha}$};
see subsection~\ref{sec:graph_types} for details.
A time series of 100 graphs is generated, with each graph comprised of 100 vertices,
and parameter {\small $\alpha$} linearly increasing from {\small $1.1$} to {\small $1.9$}.
An abnormal graph with \mbox{\small $\alpha \mbox{~=~} p_\ast  +  1.496$}
replaces the original graph at {\small $t \mbox{~=~} 50$}.
Four instances are considered, with each instance using a specific {\small $p_\ast \in \{0.25, 0.30, 0.35, 0.40 \}$}.
The value {\small $1.496$} is the original {\small $\alpha$} at the considered time point.

\begin{figure}[!tb]
  \centering
  \begin{minipage}{1\columnwidth}
    \begin{minipage}{0.025\textwidth}
      \centering
      \rotatebox{90}{\scriptsize AUC}
    \end{minipage}
    \begin{minipage}{0.31\textwidth}
      \centering
      {\scriptsize ~~~~~~TensorSplat}\\
      \vspace{0.1ex}
      \includegraphics[width=1\textwidth,height=1\textwidth]{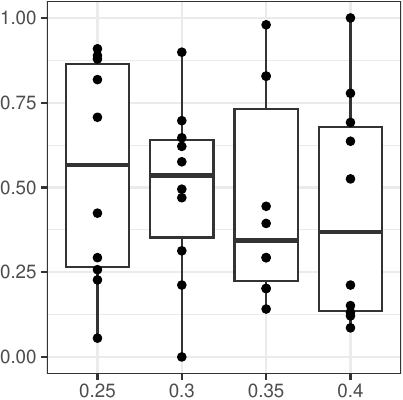}\\
      \vspace{-1ex}
      {\scriptsize ~~~~~~$p_\ast$}
    \end{minipage}
    \hfill
    \begin{minipage}{0.31\textwidth}
      \centering
      {\scriptsize ~~~~~~LAD}\\
      \vspace{0.1ex}
      \includegraphics[width=1\textwidth,height=1\textwidth]{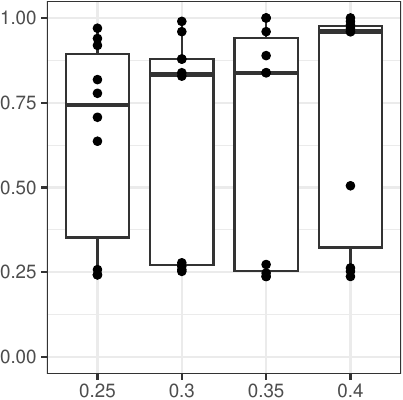}\\
      \vspace{-1ex}
      {\scriptsize ~~~~~~$p_\ast$}
    \end{minipage}
    \hfill
    \begin{minipage}{0.31\textwidth}
      \centering
      {\scriptsize ~~~~~~\textbf{proposed}}\\
      \vspace{0.1ex}
      \includegraphics[width=1\textwidth,height=1\textwidth]{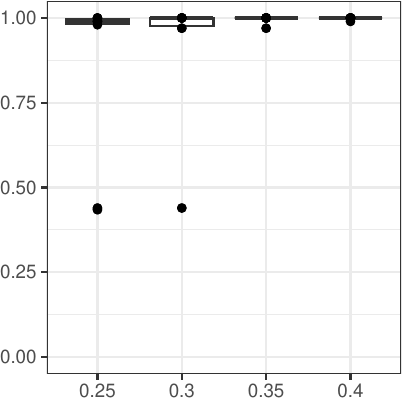}\\
      \vspace{-1ex}
      {\scriptsize ~~~~~~$p_\ast$}
    \end{minipage}
  \end{minipage}
  \vspace{-0.5ex}
  \caption
    {
    Results for experiment~3.
    As per Fig.~\ref{fig:experiment2},
    but using graphs generated according to the Barab\'asi-Albert model~\cite{barabasi}.
    In each time series, the likelihood of new vertices connecting to existing vertices is increased,
    with graph parameter {$\alpha$} linearly increasing from {$1.1$} to {$1.9$}
    (see text for details).
    In each time series, an abnormal graph is placed at $t \mbox{~=~} 50$,
    where \mbox{$\alpha \mbox{~=~} p_\ast \mbox{~+~} 1.496$},
    with {$p_\ast \in \{0.25, 0.30, 0.35, 0.40 \}$}.
    }
  \label{fig:experiment3}
  \vspace{0.5ex}
  \hrule  
\end{figure}

The results shown in Fig.~\ref{fig:experiment3} indicate that 
the proposed method achieves high overall performance,
in contrast to both TensorSplat and LAD which achieve poor performance.

\subsubsection{Experiment 4}

The Watts-Strogatz small world model is used to simulate a relatively difficult dynamic environment.
A time series of 100 graphs is generated, 
where the rewiring probability~{\small $p$} linearly increases from {\small $0.05$} to {\small $0.30$.}
Each graph has 100 vertices.
An abnormal graph with rewiring probability {\small $p_\ast + 0.1737$}
replaces the original graph at {\small $t \mbox{~=~} 50$}.
Four instances are considered, with each instance using a specific {\small $p_\ast \in \{ 0.05, 0.10, 0.15, 0.20 \}$}.
The value {\small $0.1737$} is the original rewiring probability at the considered time point.

\begin{figure}[!tb]
  \vspace{-1ex}
  \centering
  \begin{minipage}{1\columnwidth}
    \begin{minipage}{0.025\textwidth}
      \centering
      \rotatebox{90}{\scriptsize AUC}
    \end{minipage}
    \begin{minipage}{0.31\textwidth}
      \centering
      {\scriptsize ~~~~~~TensorSplat}\\
      \vspace{0.1ex}
      \includegraphics[width=1\textwidth,height=1\textwidth]{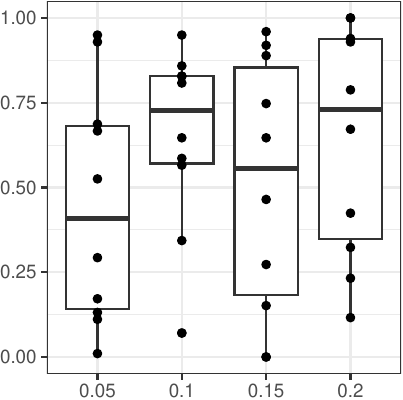}\\
      \vspace{-1ex}
      {\scriptsize ~~~~~~$p_\ast$}
    \end{minipage}
    \hfill
    \begin{minipage}{0.31\textwidth}
      \centering
      {\scriptsize ~~~~~~LAD}\\
      \vspace{0.1ex}
      \includegraphics[width=1\textwidth,height=1\textwidth]{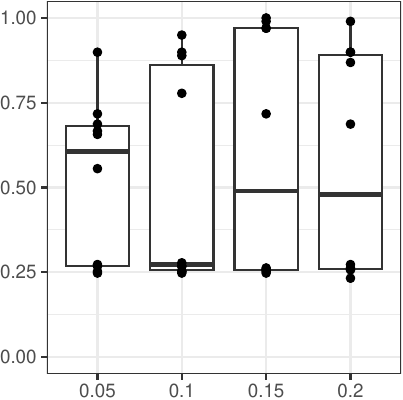}\\
      \vspace{-1ex}
      {\scriptsize ~~~~~~$p_\ast$}
    \end{minipage}
    \hfill
    \begin{minipage}{0.31\textwidth}
      \centering
      {\scriptsize ~~~~~~\textbf{proposed}}\\
      \vspace{0.1ex}
      \includegraphics[width=1\textwidth,height=1\textwidth]{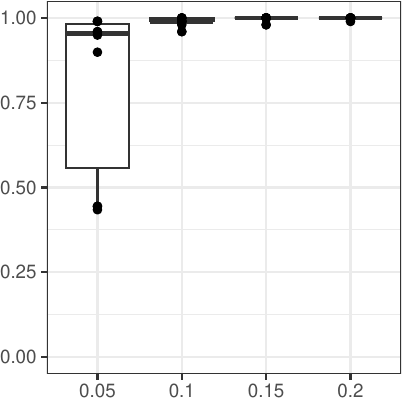}\\
      \vspace{-1ex}
      {\scriptsize ~~~~~~$p_\ast$}
    \end{minipage}
  \end{minipage}
  \vspace{-1ex}
  \caption
    {
    Results for experiment~4.
    As per Fig.~\ref{fig:experiment3},
    but using graphs generated according to the Watts-Strogatz model~\cite{Watts1998}.
    In each time series, the rewiring probability {\small $p$} linearly increases from {\small $0.05$} to {\small $0.30$.}
    In each time series, an abnormal graph is placed at $t \mbox{~=~} 50$,
    where {\small $p \mbox{~=~} p_\ast \mbox{~+~} 0.1737$},
    with {$p_\ast \in \{ 0.05, 0.10, 0.15, 0.20 \}$}.
    }
  \label{fig:experiment4}
  \vspace{0.5ex}
  \hrule  
\end{figure}

The results in Fig.~\ref{fig:experiment4} indicate that both TensorSplat and LAD struggle in this dynamic environment.
The proposed method achieves good performance for all settings of {\small $p_\ast$},
except for {\small $p_\ast \mbox{~=~} 0.05$}, while still outperforming the competing methods.

\vspace{1.5ex}
\section{Concluding Remarks}
\label{sec:conclusion}
\vspace{0.5ex}

Anomaly detection in a temporal sequence of graphs is useful in application
areas such as energy networks, transport networks and computer networks.
Anomalies can signify harmful events such as road accidents or cyber attacks.
Existing methods for detecting abnormal graphs suffer from many limitations,
including difficulties with handling variable-sized graphs and non-trivial temporal dynamics
(potentially large and/or complex but natural changes between consecutive graphs).
Furthermore, existing methods often suffer from high false positive rates.

In this work we have proposed a feature-based technique with the aim of addressing the above limitations.
Temporal dependencies are explicitly modelled via time series analysis of a large set of pertinent graph features,
followed by using residuals to remove the dependencies.
Extreme Value Theory is then used to robustly model and classify any remaining extremes,
aiming to produce low false positives rates.

Comparative evaluations on a multitude of graph instances
used as proxies for various applications
(generated via the Erd\H{o}s-R\'enyi random graph model~\cite{frieze2015introduction}, 
Barab\'asi-Albert preferential attachment model~\cite{albert2002statistical},
and Watts-Strogatz small world model~\cite{Watts1998})
show that the proposed approach obtains considerably better accuracy than
\textit{TensorSplat}~\cite{Koutra2012} and \textit{Laplacian Anomaly Detection}~\cite{Huang2020}.

Future avenues of research include
(i) an ablation study~\cite{Sheikholeslami_2021} to determine the most useful graph features,
aiding explainability~\cite{Sanderson_2023},
and
(ii)
extending the proposed method to find abnormal subgraphs in large graphs.

\newpage

\renewcommand{\baselinestretch}{0.975}\small\normalsize

\def~{\,}  

\bibliographystyle{ieee_mod}
\bibliography{references}

\end{document}